# Robust Object Tracking with Crow Search Optimized Multi-cue Particle Filter

Kapil Sharma[a], Gurjit Singh Walia[b], Ashish Kumar[a], Astitwa Saxena[c], Kuldeep Singh[d]

[a]*Delhi Technological University, Delhi, India*
[b]*Defence Research and Development Organization, Ministry of Defence, India*
[c]*Netaji Shubhas Institute of Technology, Delhi, India*
[d]*Bharat Electronics Ltd., Central Research Laboratory, Government of India, Ghaziabad, India*

**Abstract**

Particle Filter(PF) is used extensively for estimation of target Non-linear and Non-gaussian state. However, its performance suffers due to inherent problem of sample degeneracy and impoverishment. In order to address this, we propose a novel resampling method based upon Crow Search Optimization to overcome low performing particles detected as outlier. Proposed outlier detection mechanism with transductive reliability achieve faster convergence of proposed PF tracking framework. In addition, we present an adaptive fuzzy fusion model to integrate multicue extracted for each evaluated particle. Automatic boosting and suppression of particles using proposed fusion model not only enhances performance of resampling method but also achieve optimal state estimation. Performance of the proposed tracker is evaluated over 12 benchmark video sequences and compared with state-of-the-art solutions. Qualitative and quantitative results reveals that the proposed tracker not only outperforms existing solutions but also efficiently handle various tracking challenges. On average of outcome, we achieve CLE of 7.98 and F-measure of 0.734.

*Keywords:* Particle Filter, COA, object tracking, fusion model

## 1. Introduction

Object tracking is a vital research area in computer vision due to its promising applications in the field of military surveillance, augmented reality, robotics, medical imaging and human-computer interaction. It aims at keeping track of targeted object in a video sequence. Initially, target is localized either manually or through a detection algorithm and subsequently its trajectory is traced over the period of time. Keeping path of target's trajectory in a video sequence is

*Email address:* kapil@ieee.com (Kapil Sharma)

challenging due to its scale and pose variations, fast and abrupt motion, background clutters and full or partial occlusion [1].

Generally, object tracking methods can be categorized into either stochastic framework or deterministic framework. Stochastic framework include probabilistic methods in the Bayesian framework for state estimation. Under stochastic framework, linear and gaussian state estimation was addressed using kalman filter [2], extended kalman filter and unscented kalman filter [3]. Whereas non- linear and non-gaussian state can be estimated using particle filter (PF) [4], condensation filter [5] and boot strap filter [6]. On the other hand, deterministic framework viz. mean shift [7], fragment based tracker [8] and multi-stage tracker [9] employs target search in each frame to maximize similarity between the target and search space. These methods used less spatial information of the target and hence, were vulnerable to occlusion and background clutters. How- ever, stochastic methods reduced target sampling patches during tracking and were able to address the challenges occurred due to similar backgrounds and occlusion.

Under stochastic methods, PF has shown better results than other methods but its performance is severely degraded due its inherent problem of particle degeneracy and impoverishment [10]. Particle degeneracy occurs due to updation of particles with insignificant weight. These particles dispense less towards the state estimation and their usage for subsequent estimation leads to wastage of computational efforts. In order to address this, resampling techniques such as sequential resampling [11], adaptive resampling [12], dynamic resampling [13], spline resampling [14] were proposed. But implication of these methods in PF leads to sample impoverishment whereas all the particles converge to a single end leading to drastic degradation in tracker's performance. Recently, particle impoverishment is solved using nature-inspired and evolutionary optimization viz. Particle Swarm Optimization [15], Firefly Algorithm [16], Spider Monkey [17], Social Spider [18], Cuckoo Search [19], Modified Galaxy based Search [20], Bat Algorithm [21]. The exploitation and exploration characteristics of these methods introduce randomness in search space through diversification of particles. The optimal particles generated through these optimization not only achieved fast convergence rate but also enhanced the accuracy of PF frame- work. However, most of PF based tracking solution were proposed using only single feature and hence, leading to degradation in performance under dynamic environment.

Multicue based tracking framework have been discussed and reviewed mostly under deterministic methods [22]. Recently, under stochastic methods multicue based tracking framework were investigated [23], [24], [20]. In these work, multiple complementary cue are integrated to overcome individual cue limitations. In [23] author used color, texture and edge cue for the object representation and integrated them using an adaptive weighting scheme. Walia and Kapoor [24] exploited color, thermal and motion feature for object representation and integrated their scores using PCR-5 rules. In [20], authors calculated color histogram and histogram of orientation gradient (HOG) for each particle. Resampling was performed using modified galaxy based search algorithm. However, adaptive integration of multicue was not addressed. In addition, resampling methods proposed in previous work require the processing of all evaluated particles and hence, require more computational power. Hence, future tracking solutions should consider adaptive fusion of multicue to handle dynamic environmental challenges along with efficient resampling method for real time applications.

In this paper, we propose a muliticue object tracking framework based on PF with a novel nature inspired resampling method. An adaptive fuzzy based fusion model is proposed for automatic suppression and boosting of particles weights. Outlier detection mechanism is also utilized to detect low performing particles to improve the performance of resampling method. To summarize, the main contributions of our research work are as follows:

- We propose an object tracking framework in which target is represented by complementary multi-cue viz. color and texture for each particle. Color cue is sensitive to illumination variations, similar background and occlusion while texture cue can handle these variations effectively. On the other hand, texture cue fails during scale variations while color cue maintain tracker's efficiency under this challenge.

- We also propose a novel resampling method based on Crow Search Optimization (COA). COA maintain diversity in the search space by considering awareness probability. It improves performance by distributing particles in the high likelihood region to address sample degeneracy. Unimportant particles detected as outlier were resampled to reduce computational complexity of the proposed tracker.

- An outlier detection mechanism is proposed to identify unimportant particles (having negligible weight). Particle consistency in subsequent frames is exploited for detecting the particles which contribute less towards the state estimation. In addition, this mechanism improves performance of the proposed resampling technique and ensures faster convergence.

- An adaptive fuzzy fusion model is proposed for integration of multi cues. This model resolve conflict among the particles and takes decision on the basis of fuzzy inference rule. It either boost or suppress the particles for their optimal contribution in the state estimation. It creates clear decision boundaries between significant particles and low performing particles. Model is made adaptive through a transductive reliability calculated through correlation among the particles.

The rest of this paper is organized as follows. In Section II, we review the literatures closely related to our work. Section III discusses the architecture of the proposed multicue object tracking framework. Detailed design of resampling model and fusion model is also presented. Core design of proposed tracker is highlighted further in section IV. Qualitative and quantitative results along- with their comparison with existing methods is presented in section V. Finally, concluding remarks and future directions are drawn in section VI.

## 2. Related Work

Object tracking under PF framework has been extensively investigated due to its potential applications in video surveillance, military and robotics. However, PF inherent problem viz. sample impoverishment and degeneracy limit its performance of object tracking. Many solutions have been proposed to address these problems and reviewed in [10]. In this section, we examine the closely related work of contemporary solutions for object tracking under PF framework. In [16], author used firefly algorithm (FA) before resampling to improve particle distribution. FA converged particles rapidly to optimal solution by changing their positions. Use of optimization generate significant particles which leads to accurate target state estimation. However, performance of this method de- graded when FA trapped in local minimum. Walia and Kapoor [19] presented a resampling method based on improved cuckoo search (ICS) with levy flight for object tracking under PF framework. ICS abandoned the particles with lesser weight to avoid the local minimum trap. However, optimal selection of step size for levy flight was not addressed. In [17], authors utilized spider mon- key optimization as resampling method in PF based tracking framework. The particle's position was updated on the basis of experience of local and global leader. Use of optimization improved quality of particles and converge them in high likelihood region. Resampling method based on immune genetic algorithm (IGA) for object tracking under PF framework was discussed [25]. IGA used crossover and mutation process to generate new particles in search region. IGA's regulatory mechanism viz. promotion and suppression ensured particles diversity. In [26] author used Pearson correlation coefficient as resampling under PF framework. Similarly, In [27] authors explored resampling method based on backtracking search optimization (BSA). But both of these methods were not evaluated on video sequences. Ahmadi and Salari [18] applied social spider optimization (SSO) as resampling method for tracking framework based on PF. SSO grouped the particles as male and female to increase randomness in search space. If target was lost, optimization ensured diversity by forcing the redistribution of particles. However, most of the methods under PF framework consider only single cue for particle evaluation. However, single cue is not sufficient to handle dynamic environments. Hence, particles need to be evaluated over multicue and subsequently fused for compensating various tracking challenges.

In [15], authors exploited color histogram and HOG for evaluation of each particle and PSO is applied as resampling technique. But this method is computationally complex due to multilevel evaluation of each particles with multiple cues. In [20], authors used color histogram and HOG cue with resampling method based on modified galaxy based search. Adaptive appearance model with multiple parameters were used at each time step to detect and handle occlusion. Due to this, method was considerably slow. Similarly, Sardari and Moghaddam [28] used color and motion cue for likelihood calculation of particles. The resampling method based on modified galaxy based search algorithm was proposed to obtain optimum state of the object. In [24], authors exploited color, thermal and motion cue for each particle evaluation. The particles were fused using Dezert-Smarandache Theory. Generally, resampling mechanism for PF consider all the particles to ensure optimal state estimation. This needs high computational power for desired convergence. Methodology to relocate the lost target has not been explored much. Moreover, adaptive fusion of particles in multi cue object tracking based on PF is limitedly addressed.

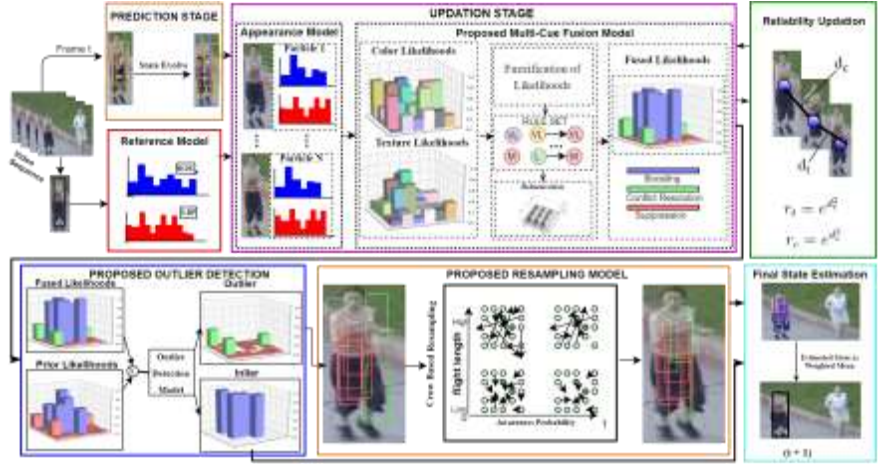

Figure 1: Overview of the proposed Tracker. *N* particles are evolve using multi component state model. Predicted particles are updated through multicue evaluation and adaptive fusion of evaluated particles. Detected outlier are resampled to overcome sample degeneracy. Targeted state is determined as weighted mean of *N* particles.

Review of closely related literatures indicate that future research in multicue object tracking based on PF should consider adaptive fusion of particles for efficient tracking. In addition, unimportant particles needs to be detected not only to reduce computation but also for efficient resampling. The next section will detail architecture of the proposed tracker.

## 3. Architecture of Proposed Tracker

The Proposed tracker is developed under PF based probabilistic tracking framework with multicue evaluation for each and every particle. Extracted color cue ($H_C$) and texture cue ($H_T$) for each evaluated particle are integrated using the proposed fuzzy based adaptive fusion model. Fusion model is made adaptive through online calculation of transductive reliability metric for each cue. The particle level multicue fusion is performed considering boosting of good particles and simultaneously suppressing of the low performing particles. In each frame, low performing particles are detected using the proposed outlier detection mechanism. These particles are relocated to their optimum position using the proposed resampling method based on Crow Search Optimization (COA) [29]. Architecture of the proposed tracker is depicted in Fig 1. As shown, at $k = 1$, the target is segmented using Gaussian Mixture Model (GMM) [30] and *N* particles are initialized around the detected centroid of the target. These initialized particles are evolved using the Multi component state model to obtain the predicted particles. Each predicted particle is evaluated over two complementary cues and weights are assigned through comparison of extracted features with the corresponding reference model. Further, these weights are integrated using the multicue fuzzy fusion model to obtain fused weight $w_t$ for each evaluated particles. Based upon their fused weights particles are classified into outlier ($O$) and inlier ($I$) using the proposed outlier detection mechanism which is based on Markov process. Only the outliers $O$ are processed through the proposed resampling method to relocate them in high likelihood area. Resampling mechanism considers outlier as initial population for COA. The diversity of the particles is maintained through the awareness probability. On the other hand, search region is controlled through the flight length to obtain the optimal resampled particles. High convergence rate of COA achieves real time realization of the proposed tracker. At each time step, transductive reliabilities $r_c$ and $r_t$ associated with the cues are evaluated to discount the particle in concurrence with the cue performance. These reliability ensures quick adaptation of the pro- posed tracker to the dynamic environment. Finally, target state is determined as weighted sum of inlier as well as resampled outlier particles. The detailed design of the proposed tracker is follows in turn.

*3.1. Multicue Particle Filter*

Proposed tracker is designed using Particle Filter [6] which is a probabilistic recursive Bayesian framework used for estimation of Non-linear and Non- Gaussian state. Particle filter has been exploited for multicue tracking frame- work to precisely localize the target under dynamic environment. Aim is to approximate PDF $p(S_k|w_k)$ of the target state given its prior PDF $p(S_{k-1}|w_{k-1})$, where $S_k$ and $S_{k-1}$ are state vectors represented by $N$ particles and $w_k$ and $w_{k-1}$ are set of measurements at time instant $k$ and $k-1$.

In the proposed model, we consider state vector for each particle as $S_k = (X_k, X_k^j, Y_k, Y_k^j, r_k, \theta_k)$. $X_k$ and $Y_k$ are cartesian coordinates of the center of bounding box, $X_k^j$ and $Y_k^j$ are velocities in respective direction, $r_k$ and $\theta_k$ are rotation and scaling components of the particle. Measurement is obtained for each particle using complementary multicue viz. color and texture extraction and their score fusion using proposed multicue fusion model. Particles are evolved using the state model as Eq.(1).

$$S_k = f_k(S_{k-1}, r_{k-1}) \qquad (1)$$

Where, $r_{k-1}$ is zero mean white noise and $f_k: R^p \; R^d \; R^p$ is a multicomponent transition function which includes constant velocity model and random walk model [23].

The PDF of target state $S_k$ i.e. $p(S_k|w_k)$ is obtained recursively into two stages viz. prediction and update. The initial PDF of the target $p(S_{k-1}|w_{k-1})$ at time $k = 2$ is manually obtained through random placement of particles. In the subsequent estimate prior PDF of the target at time instant $k$ is obtained using

Eq.(2).

$$p(S_k|w_{k-1}) = \int p(S_k|S_{k-1})p(S_{k-1}|w_{k-1})dS_{k-1} \quad (2)$$

Where, multicomponent state model is used to evolve the particles to obtain predicted particle represented PDF $p(S_k|S_{k-1})$. Further, the predicted particles representing PDF $p(S_k|S_{k-1})$ are updated by assignment of multicue measurement and fusion as discussed in Section B. The updation of these particles is performed using Bayes rule using Eq. (3).

$$p(S_k|w_k) = \frac{p(w_k|S_k)p(S_k|w_{k-1})}{p(w_k|w_{k-1})} \quad (3)$$

In Eq(3) normalizing denominator factor $p(w_k|w_{k-1})$ is obtained as:

$$p(w_k|w_{k-1}) = \int p(w_k|S_k)p(S_k|w_{k-1})dS_k \quad (4)$$

Where, $p(w_k|S_k)$ is obtained through multicue evaluation of each particle and their subsequent fusion using proposed mulitcue fusion model.

Posterior PDF $p(S_k|w_k)$ is represented by $N$ updated particles and used to determine the state of the target as their weighted mean. In the iterative process of particle filter these updated particles are used as prior initialization for the next stage estimation. Details of the proposed tracker updation model is discussed in the next section.

*3.2. Proposed Updation Model*

Proposed tracker is realized through online detection of target through GMM [30] where $N$ particles are initialized. Each particle state $S_k$ is evolved through multi component state model to achieve predicted particles. Particles updation consists of mainly: a) multicue evaluation and b) adaptive fusion of likelihoods. For this, each particle is evaluated over two complementary cues viz. color and texture. Color cue is invariant to scale variations and occlusion but sensitive to illumination variations and background clutters. On the other hand, texture cue is robust to illumination variations and background clutters. Hence, complementary color and texture cue compensate for each other during tracking process.

The color cue is calculated using RGB color histogram model. Color histogram ($H_c$) for each $t^{th}$ particle with pixel location p ($a_n$, $b_n$), is determined by Eq. (5).

$$H_c = Z \sum_{t=1}^{N} B(a_i, b_i), c = 1,2 \ldots M_b \quad (5)$$

where, $B(.)$ represents binning function that assign pixel ($a_i$, $b_i$) to one of the $M_b$ histogram bins and normalizing factor $Z$ is calculated as $\sum_{c=1}^{M_b} H_c = 1$.

Unlike color cue, texture cue is used to extract low level features from the image. It is determined for each particle by considering rotation invariant Local

Binary Pattern (LBP) [31]. The LBP of $t^{th}$ particle is calculated from central pixel $(a_c, b_c)$ with radius $z$ and with equally spaced pixels $p$. The uniform pattern in image region used multi-resolution gray scale rotation for separating target texture information. It is represented as $U(LBP_{z,p}((a_c, b_c))$ and determined using Eq. (7).

$$LBP_{(z,p)}(a_c, b_c) = \begin{cases} \sum_{j=0}^{p-1} T\left(\left(u_j(a_j, b_j) - u_c(a_c, b_c)\right)\right), & if\ U(LBP_{z,j}(a_c, b_c)) \leq 2 \\ j+1, & otherwise \end{cases} \quad (6)$$

$$U(LBP_{(z,p)}(a_c, b_c)) = |T\left(u_{j-1}(a_{j-1}, b_{j-1}) - u_c(a_c, b_c)\right) - T(u_0(a_0, b_0) - u_c(a_c, b_c))$$
$$+ \sum_{j=0}^{p-1} |T\left(u_j(a_j, b_j) - u_c(a_c, b_c)\right) - T\left(u_{j-1}(a_{j-1}, b_{j-1}) - u_c(a_c, b_c)\right)| \quad (7)$$

Using this LBP operator, normalized histogram are determined for each particle by mapping uniformity measure of every pixel and represented as $H^t$. Further, $_T$ weights are assigned to each particle through estimation of similarity between predicted particles histogram and corresponding references histograms. This is achieved through Bhattacharyya distance [32]. For this, Bhattacharyya coefficient is determined using Eq. (8).

$$\beta_{t,i}(H_i^R, H_i^t) = \sum_{r=1}^{N} \sqrt{(H_i^R)^r \times (H_i^t)^r} \quad (8)$$

Where, $H_i^R$ represents reference histogram, $i \in C, T$ and $H^t$ denotes the histogram of considered cue for $t^{th}$ particle. Using Eq. (8) Bhattacharyya distance is calculated for color and texture cue as illustrated in Eq. (9).

$$BD_{t,i}(H_i^R, H_i^t) = \sqrt{1 - \beta_{t,i}(H_i^R, H_i^t)} \quad (9)$$

Using this, likelihoods for each particle are determined using Eq. (10).

$$\alpha_{t,i}(H_i^R, H_i^t) \propto e^{\left(-\frac{(BD_{t,i}(H_i^R, H_i^t))^2}{2\sigma_i^2}\right)} \quad (10)$$

Where $\sigma_i$ denotes standard deviation and $I \in (C, T)$.

The color and texture likelihoods so calculated are subjected to the proposed adaptive multicue fusion model. The proposed model is designed considering boosting of concordant particles, suppressing of discordant particles and on-line conflict resolving among the conflicting particles. In order to achieve quick adaptation of tracker, individual likelihoods are discounted with the transductive reliability of the respective cue as determined in the previous frame. For this, particle likelihoods $\alpha_{t,i}$ are multiplied with the corresponding cue reliability as depicted in Eq (11).

$$\gamma_{t,i} = r_i \times \alpha_{t,i}(H_i^R, H_i^t) \tag{11}$$

Where, $r_i$ is transductive cue reliability determined in the previous frame and $i \in C, T$ ). Proposed fuzzy fusion model achieves precise decision boundary among high performing particles and low performing particles under complex environment. Fuzzy logic can induce human being inferences in the fused model [33]. It was reported that human sensing system consist of multi cue framework with online adaptation to dynamic environment [34]. Considering this, fuzzy set is defined as: S ={V S, S, M, L, V L}representing Very Small, Small, Medium, Large and Very Large likelihood value for the particle. Membership value for the evaluated particle $\gamma_{t,i}$ are determined in each fuzzy set element using Eqs. (12-16):

$$\delta_{VS}(x) = \frac{1}{1+e^{-c_{VS}(x-f_{VS})}} \tag{12}$$

$$\delta_S(x) = e^{\frac{-(x-d_S)^2}{f_S^2}} \tag{13}$$

$$\delta_M(x) = e^{\frac{-(x-d_M)^2}{f_M^2}} \tag{14}$$

$$\delta_L(x) = e^{\frac{-(x-d_L)^2}{f_L^2}} \tag{15}$$

$$\delta_{VL}(x) = \frac{1}{1+e^{-c_{VL}(x-f_{VL})}} \tag{16}$$

Where, $c_{V\,S} = 80, f_{V\,S} = 0.15, d_S = 0.3, f_S = 0.12, d_M = 0.5, f_M = 0.12, d_L = 0.7, f_L = 0.12, c_{V\,L} = 80$ and $f_{V\,L} = 0.15$ and $x \in (\gamma_{t,i}, w_t), i \in (C, T)$.

The functional mapping $R_{m,n}$ between the color and the texture likelihood is presented in the Table 1, where *m* and *n* represent the fuzzy set value assigned to each evaluated particle. This mapping ensures boosting of high performing particles and suppression of low performing particles, creating a precise decision boundary for the proposed outlier detection mechanism.

| m/n | V S | S  | M  | L   | V L |
|-----|-----|----|----|-----|-----|
| V S | V S | V S| V S| M   | L   |
| S   | V S | V S| S  | M   | L   |
| M   | V S | S  | M  | M   | L   |
| L   | M   | M  | M  | V L | V L |
| V L | L   | L  | L  | V L | V L |

Table 1: Fuzzy inference Model ($R_{m,n}$)

Further to achieve crisp fused value of particle likelihood, defuzzification using Center of Gravity (COG) method has been considered for obtaining optimal results. The crisp COG value for a pair of fuzzy set elements *m, n* is defined as:

$$COG_{R_{m,n}} = \frac{\int_0^1 \delta_{R_{m,n}(x)} x \, dx}{\int_0^1 \delta_{R_{m,n}(x)} \, dx} \tag{17}$$

Where $R_{m,n}$ is determined using Table 1 and $\xi$ ($\gamma_{t,i}$, $w_t$). The weighted mean, Eq. (18), of COG values being determined over pair ($m, \in n$) $\&$ $S$ and used for estimation of final fused weight for each evaluated particle.

$$w_t = \frac{\sum_m \sum_n \pi_{m,n}(\gamma_{t,C}, \gamma_{t,T}) \times COG_{R_{m,n}}}{\sum_m \sum_n \pi_{m,n}(\gamma_{t,C}, \gamma_{t,T})} \quad (18)$$

Where, $\pi_{m,n}$ is fuzzy control rule calculated using Eq. (19).

$$\pi_{m,n}(\gamma_{t,C}, \gamma_{t,T}) = \min(\delta_m(\gamma_{t,C}), \delta_n(\gamma_{t,T})) \quad (19)$$

*3.3. Proposed Resampling Method*

Proposed multicue fusion model creates precise decision boundaries between low performing particles and high performing particles. Keeping low performing particle for subsequent iteration not only requires high computation but also impairs tracker accuracy. This problem of sample degeneracy has been solved using conventional techniques like adaptive resampling [12], sequential resampling [11], cuckoo search [19]. However, most of the resampling methods consider all the particles for resampling process which requires high computational requirement. In order to realize the tracker in real time scenario, we determine the recurrently low performing particles as outliers which are subjected to the proposed resampling method. We assume that outliers are those particles which are either drifted off the target or highly affected by the dynamic environment. Generally, environments affect the target particles in consequent frames. This has been exploited by assuming that individual particle state estimation as a Markov process and helps us in design of proposed outlier detection mechanism. Proposed outlier detection mechanism is designed by considering particles performance in previous frame along with their current estimation. The particles likelihood for the current state ($w^k$), $t$ 1...$N$ are fused with their previously estimated likelihoods ($w^{k-1}$). These likelihoods for each particle are adaptively fused using the proposed multicue fusion model as discussed in Section B. For this, we consider the fuzzy set $S$ for the two likelihood value for each particle and membership values ($\delta_m(.)$) are determined using Eqs. (12-16). Further, final score for the $t_{th}$ particle for its classification as outlier or inlier is calculated using Eq. (20).

$$\kappa_t = \frac{\sum_m \sum_n \pi_{m,n}(w_t^k, w_t^{k-1}) \times COG_{R_{m,n}}}{\sum_m \sum_n \pi_{m,n}(w_t^k, w_t^{k-1})} \quad (20)$$

Where, $\pi_{m,n}$ is fuzzy control rule calculated using Eq. (21) and used for weighing the COG values.

$$\pi_{m,n}(w_t^k, w_t^{k-1}) = \min(\delta_m(w_t^k), \delta_n(w_t^{k-1})) \quad (21)$$

Using the fused score for each particle, outlier and inlier are classified by comparing their fused score with predefined threshold value $\kappa_{th}$. The particles are classified as outliers using Eq. (22) and as inliers using Eq. (23)

$$O = \{S_k^t, \forall\ t\ where\ \kappa_t \leq \kappa_{th}\} \quad (22)$$

$$I = \{S_k^t, \forall\ t\ where\ \kappa_t > \kappa_{th}\} \quad (23)$$

Precise decision boundary determined through proposed outlier detection mechanism aids proposed resampling mechanism to eliminate particle degeneracy. Only the particles classified as outliers are resampled using the proposed resampling technique. The resampling technique is based on crow search optimisation which diversifies the outliers in the search space. The crow search optimization is a meta-heuristic optimization approach proposed in [29]. The outlier particle set $O_k$ obtained through Eq. (22) are initialized as the crow population. Position of the crow population is updated as shown in Eq. 24.

$$O_v = \begin{cases} O_v + r \times fl \times (M_v^r - O_v), & r \geq AP \\ random\ redistribution, & otherwise \end{cases} \quad (24)$$

where $r \in [0, 1]$, $v \in (X, Y)$, $fl$ represents the flight length and $AP$ denotes the awareness probability and $M_{k,v}^r$ represents the global best location of each crow. This iterative process is performed for $J$ iterations to achieve the optimal solution. Finally, the optimally localized outliers with updated weights are used for determining the target state. Proposed tracker state estimation is depicted

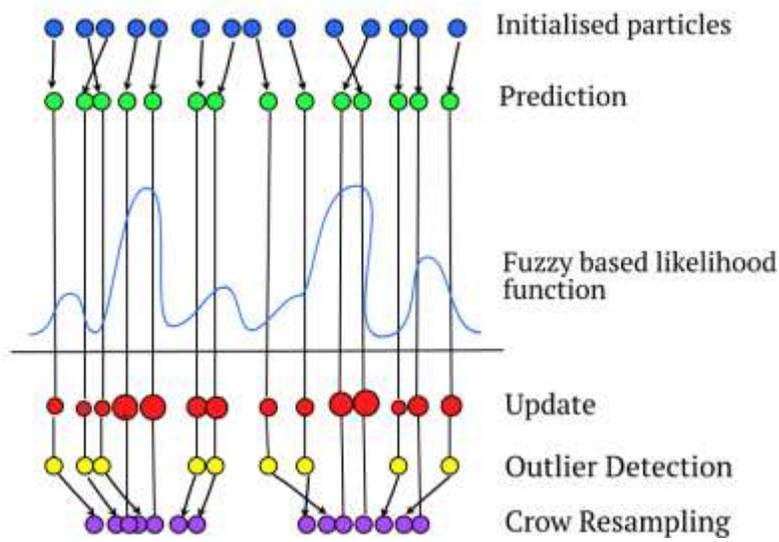

Figure 2: Optimal particle localization using resampling process

in Fig. 2 where particle marked as outlier are detected and resampled. After

**Algorithm 1:** Proposed Tracker

```
 1  Function Multicue Tracker():
 2      for each frame k do
 3          if k==1 then
 4              Intialise N particles on the detected target
 5              r_C, r_T ← 1,1
 6          else
 7              S_k = f_k(S_{k-1}, w_{k-1})
 8              Obtain α_{t,i} using Eq. (10)
 9              γ_{t,i} ← r_i × α_{t,i}, i ∈ (C, T)
10              Calculate w_t using Eq. (19)
11              Obtain κ_t using Eq. (20)
12              I, O ← φ
13              for t in S_k do
14                  if κ_t > κ_{th} then
15                      I = I ∪ S_t^k
16                  else
17                      O = O ∪ S_t^k
18                  end
19              end
20              for j from 1 to J do
21                  for c from 1 to |O| do
22                      r ← U(0,1)
23                      R ← Uint(J)
24                      if r ≥ AP then
25                          O_{v,c} ← O_{v,c} + r×fl×(M_v^R - O_{v,c})
26                      else
27                          O_{v,c} ← O_{v,c} + N(0, 1)
28                      end
29                      Update M_v^c
30                  end
31              end
32              Obtain updated w_t using Eq. (19)
33              Calculate S^F, S^F, S^F using Eqs. (25-27)
34              calculate d_C = σ(S_k^F, S_k^R)
35              calculate d_T = σ(S_k^F, S_k^T)
36              Obtain r_C, r_T using Eqs. (28-29)
37          end
38      end
39      return S_1, S_2, ..., S_k
40  Mt
```

resampling the outliers are positioned appropriately for improvement in their performance for state estimation. The final state estimation is described in the following section.

*3.4. State Estimation and Reliability Updation*

The final state of the object ($S_k^F$) is estimated by using the updated fused likelihoods $w_t$ as weights for inlier states ($I_{k,v}$) and resampled outlier states ($O_{k,v}$). This is depicted in Eq. (25).

$$S_k^F = \frac{\sum_u w_u \times O^u + \sum_v w_v \times I^v}{\sum_{t=1}^N w_t} \quad (25)$$

where $|u \cup v| = N$

For quick adaptation of the proposed tracker against environment challenges, reliability values are calculated for each cue individually. For this, the state is estimated using color and texture likelihoods and depicted in Eqs. (26-27).

$$S_k^R = \frac{\sum_u \gamma_{u,C} \times O^u + \sum_v \gamma_{v,C} \times I^v}{\sum_{t=1}^N \gamma_{v,C}} \quad (26)$$

$$S_k^T = \frac{\sum_u \gamma_{u,T} \times O^u + \sum_v \gamma_{v,T} \times I^v}{\sum_{t=1}^N \gamma_{v,T}} \quad (27)$$

The reliability values are obtained through Eqs. (28,29), which are further passed on to the next frame iteration.

$$r_C \propto e^{-d_C} \quad (28)$$

$$r_T \propto e^{-d_T} \quad (29)$$

where $d_C$, $d_T$ are the Euclidean distance between the final state and the state estimated for color and texture cue respectively and is given by Eqs. (30-31)

$$d_C = \sigma(S_k^F, S_k^R) \quad (30)$$
$$d_T = \sigma(S_k^F, S_k^T) \quad (31)$$

The pseudocode of the proposed tracker is depicted in Algorithm 1. Updated inliers ($I^v$) and Outliers ($O^u$) are used as a prior for next state estimation. These particles are further evolved using the state model Eq.(1). Predicted particles are updated using the multicue framework as discussed in Section B. In addition, transductive reliabilities estimated at ($k$) state are used for discounting the particles weights at ($k + 1$) state. This iterative process will continue till the detected target is present in the scene. Once the target is lost, the automatic detection mechanism using GMM will redetect the new target and the particles are re-initialized in the iterative process. Details of experimental validation of the proposed tracker along with the comparison with state-of-the-art is presented in the next section.

## 4. Experimental Validation

In this section, we demonstrated experimental validation of the proposed method and compare the results with 9 other state-of-the-art trackers viz. ASLA [35], CT [36], MIL [37], IVT [38], FRAG [8], WMIL [39], PF [6], PF-PSO [15] and PF-BSA [27] . In order to gauge robustness and reliability of the proposed method, we have evaluated our tracker over 12 benchmarked video sequences, encompassing various tracking challenges such as illumination variations, scale variations, full or partial occlusion, background clutters, object's deformation and rotation. Qualitative and quantitative analysis of results is performed using standard performance metrics. Proposed algorithm is implemented in Python2 and executed on a 2.4 GHz quad core processor with 6 GB RAM. Proposed tracker is initialized with $N = 49$ particles on pre-detected target using GMM [30] segmentation technique. Probabilistic nature of PF is handled by iterating the algorithm 10 times and taking the mean value of the results. 1-D and 2-D tracking work of Bhateja et al. [27] has been extended for video sequences to compare its performance with the proposed method. In the next section, the qualitative analysis of the proposed method is demonstrated.

*4.1. Qualitative analysis*

*4.1.1. Illumination and Scale variations*

The challenging video sequence *CarScale* frames under illumination and scale variations have been illustrated in Fig. 3 (a). From frames #180 to #240, under prominent variation in the target's scale, the trackers viz ASLA, CT, MIL, WMIL and IVT totally deflected from the target. Due to their appearance model, these trackers fail to handle the large-scale variations of the target but our tracker can efficiently track the target. Proposed multicue fuzzy fusion model suppresses the low performing particles during the large-scale variations and hence, maintains tracker's efficiency. In Fig. 3(b) sample frames for sequence *Dog* are shown. At #126 frame, when the targeted object encounters scale variations PF-PSO, IVT and PF drifted away while ours successfully tracked the marked object. The representative frames for the sequence *Human8* are illustrated in Fig. 3(c), target has significant change in illumination from frame #100 to #124. Only the proposed tracker keeps track of the target. This is mainly attributed to multi component state model where scaling factor is included in the state vector. There is a sudden change in illumination at frames #41, #132 and #229 for sequence *Singer2* as depicted in Fig. 3(d). Discriminative trackers CT and MIL are not able to handle the drastic change in illumination. These trackers considered Haar like features for object representation which are prone to illumination variations. However, the proposed tracker is robust to illumination variations as it considers complementary cues viz. color and texture. When RGB color histogram performance degrades due to illumination and scale variations, the scale invariant LBP texture histogram will compensate. In addition, the particles which are affected by the illumination and scale variations are suppressed by the proposed multicue fuzzy fusion model and hence, enhances the tracking accuracy.

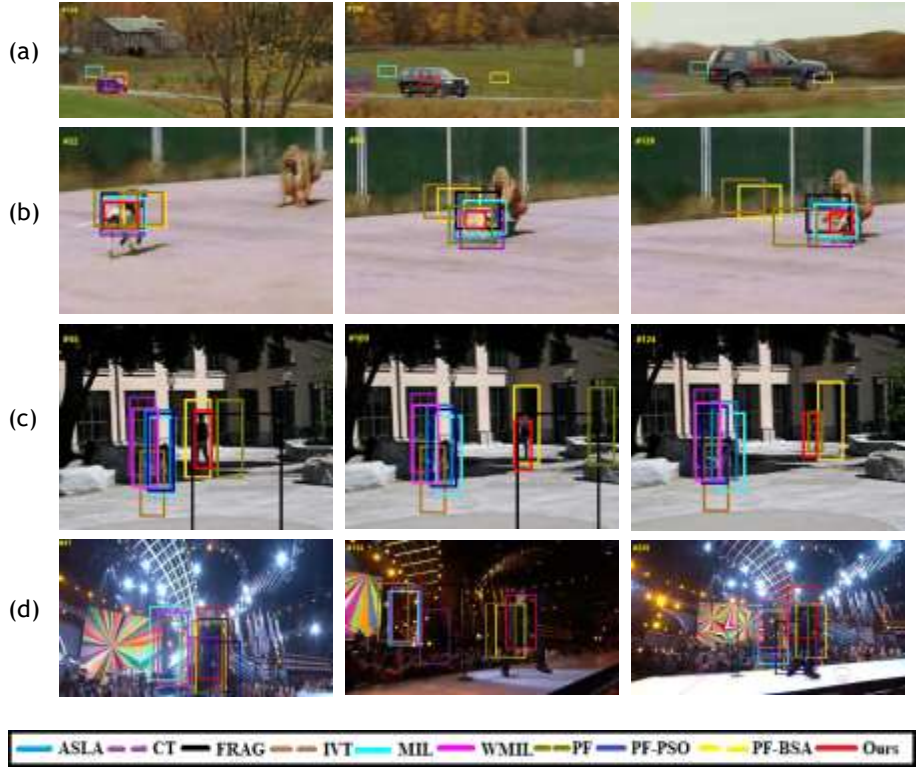

Figure 3: Critical frames with Illumination and Scale variations video sequences: a) *CarScale* b) *Dog* c) *Human8* d) *Singer2*

*4.1.2. Full or Partial Occlusion*

The sample frames for video sequences considered under this challenge are presented in Fig.4. The *Bolt* sequence has several full and partial occlusions as shown in Fig. 4(a). ASLA, CT, MIL, WMIL, PF-PSO, PF-BSA and IVT fail to track the target when it undergoes full occlusion at frame #126 and #275 while ours perform comparatively better. Fig 4(b) depicts partial occlusions of the target at frame #190. CT, MIL and Frag has shown unsatisfied performance under partial occlusion. On the other hand, our tracker with adaptive multicue fuzzy fusion model boost the high performing particles during occlusion and hence, prevents drift. At frame #70 target has full occlusion by the pillar as shown in Fig. 4(c). The trackers viz. ASLA, MIL, WMIL, CT, FRAG, IVT and PF are not sufficiently able to handle this challenge as they are not able to alleviate from drift. Whilst ours perform effectively under this challenge. Critical frames for *Subway* sequence are shown in Fig. 4(d). At frame #90 ASLA and WMIL lost the target when it is partially occluded by the other person. Ours along with MIL and WMIL has shown relatively good performance in comparison to the other trackers. Proposed tracker performance improvement under full or partial occlusion is mainly attributed to the outlier detection mechanism and the proposed resampling technique. The particles affected due to occlusion are identified by the outlier detection mechanism. The proposed resampling technique diversified these particles in the search space in order to improve their placement. This helps in maintaining the tracker's efficiency under full or partial occlusion.

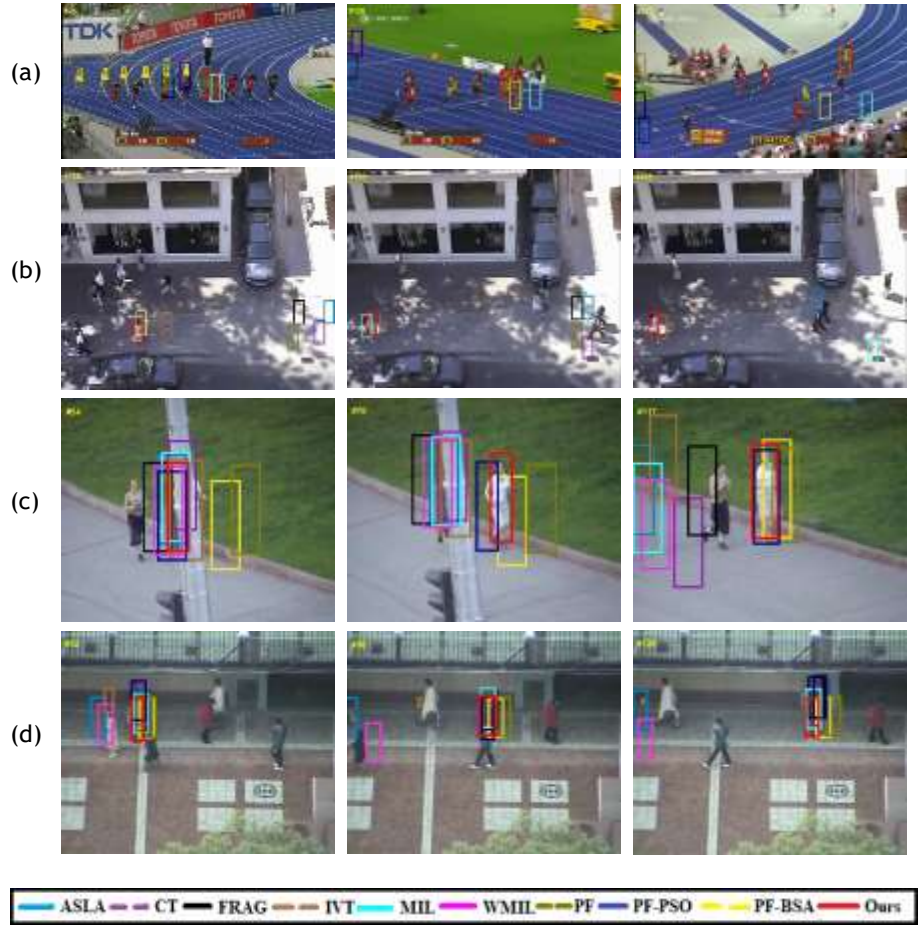

Figure 4: Critical frames with Full or partial Occlusion for video sequences: a) *Bolt* b) *Crowds* c) *Jogging-2* d) *Subway*

*4.1.3. Background clutters*

We have considered challenging video sequences under this challenge and certain frames are presented in Fig. 5. In *Walking* video sequence at frame #125 PF drift away from the target as shown in Fig 5(a). PF-BSA and ASLA also drift away from the target as they have no mechanism to relocate the target but ours, MIL and FRAG successfully track the target. In Fig. 5(b) critical frames for *MountainBike* sequences are shown. MIL, WMIL, CT, FRAG, PF and PF-BSA are inefficient to handle the similar background. However, ours, ASLA and MIL is able to track the target. In our tracker, the outlier detection mechanism identified the unimportant particles due to similar background and hence, prevents false updation of particles in our tracker. *Crossing* critical frames are presented in Fig. 5(c). At frame #94 after the car passes in front of the target, ASLA, FRAG, PF-BSA and PF deflected away from the target. On the other hand, our tracker has shown substantially good performance in the sequence as proposed adaptive multicue fuzzy fusion supports boosting of good particles during the background clutters. The frames for sequence *Couple* are shown in Fig. 5(d). After frame # 90 when the target passes by the similar car in the background then only ours with FRAG able to track the target efficiently. The tranductive reliability values of the color and texture prevents the drift of target during background clutters and the redetection strategy also supports in enhancing the tracking accuracy.

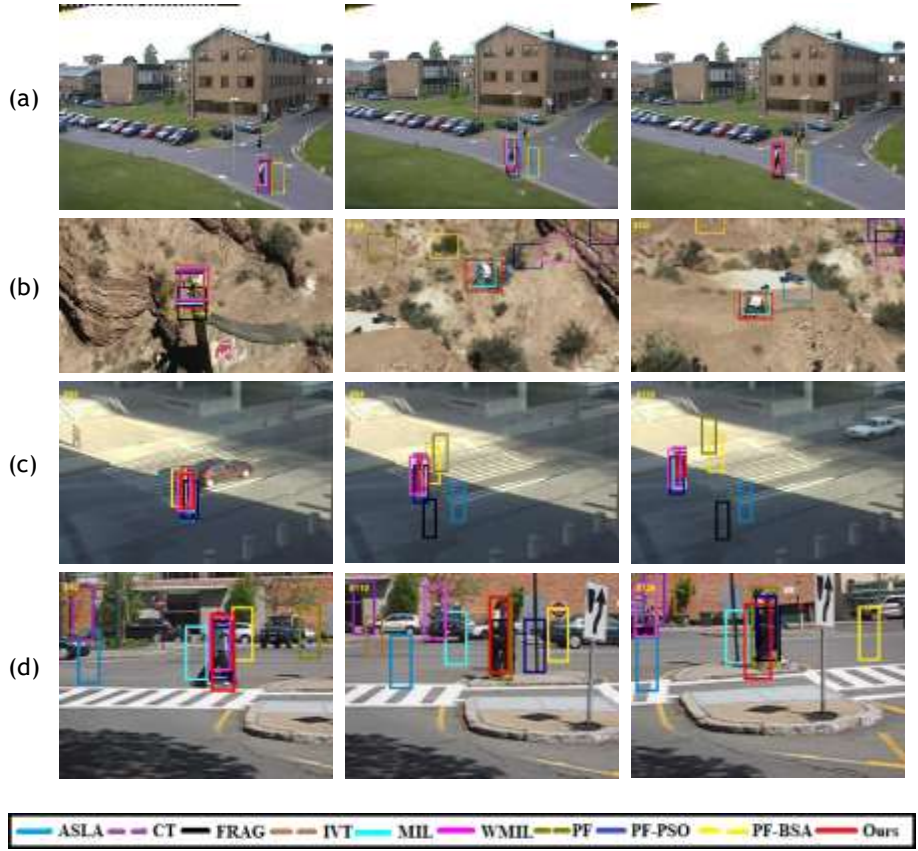

Figure 5: Critical frames with background clutters for video sequences: a) *Walking* b) *MountainBike* c) *Crossing* d) *Couple*

| Video Sequence | ASLA | CT | FRAG | IVT | MIL | WMIL | PF | PF-PSO | PF-BSA | Ours |
|---|---|---|---|---|---|---|---|---|---|---|
| *CarScale* | 79.80 | 75.14 | 31.21 | 11.58 | 70.97 | 70.33 | 40.26 | 38.08 | 96.86 | 7.26 |
| *Dog* | 9.21 | 8.33 | - | 46.81 | 13.42 | 25.13 | 25.44 | 13.51 | 37.26 | 8.98 |
| *Human8* | 65.56 | 82.65 | 55.65 | 90.01 | 62.77 | 79.26 | 32.70 | 71.90 | 30.12 | 5.96 |
| *Singer2* | 193.91 | 121.21 | 88.44 | 171.09 | 161.69 | 160.98 | 34.26 | 140.23 | 56.49 | 13.76 |
| *Bolt* | 379.76 | 376.24 | 333.81 | 367.76 | 383.71 | 370.01 | 68.07 | 91.14 | 64.09 | 9.39 |
| *crowds* | 295.40 | 380.72 | - | 11.04 | 261.56 | 249.34 | 110.91 | 10.49 | 146.51 | 5.83 |
| *Jogging2* | 143.90 | 116.98 | 37.42 | 132.33 | 134.43 | 141.62 | 16.05 | 13.48 | 24.68 | 7.06 |
| *Subway* | 145.04 | 11.58 | 8.50 | 117.19 | 6.70 | 136.73 | 9.71 | 11.34 | 5.73 | 4.06 |
| *Walking* | 247.76 | 8.55 | 9.54 | 2.30 | 4.38 | 8.63 | 81.70 | 15.72 | 45.08 | 8.15 |
| *Mountainbike* | 21.73 | 192.44 | 206.90 | 7.21 | 7.78 | 107.11 | 165.75 | 94.76 | 85.63 | 8.31 |
| *Crossing* | 33.41 | 5.37 | 38.50 | 3.18 | 3.16 | 5.12 | 22.66 | 12.23 | 16.19 | 3.19 |
| *Couple* | 96.23 | 79.30 | 8.94 | 100.50 | 27.68 | 39.78 | 38.81 | 65.10 | 42.78 | 13.77 |
| **Average** | 142.64 | 121.54 | 81.89 | 88.42 | 94.85 | 116.17 | 53.86 | 48.16 | 54.28 | 7.98 |

Table 2: Centre Location Error (in pixels). The best results are shown in red, blue and green respectively.

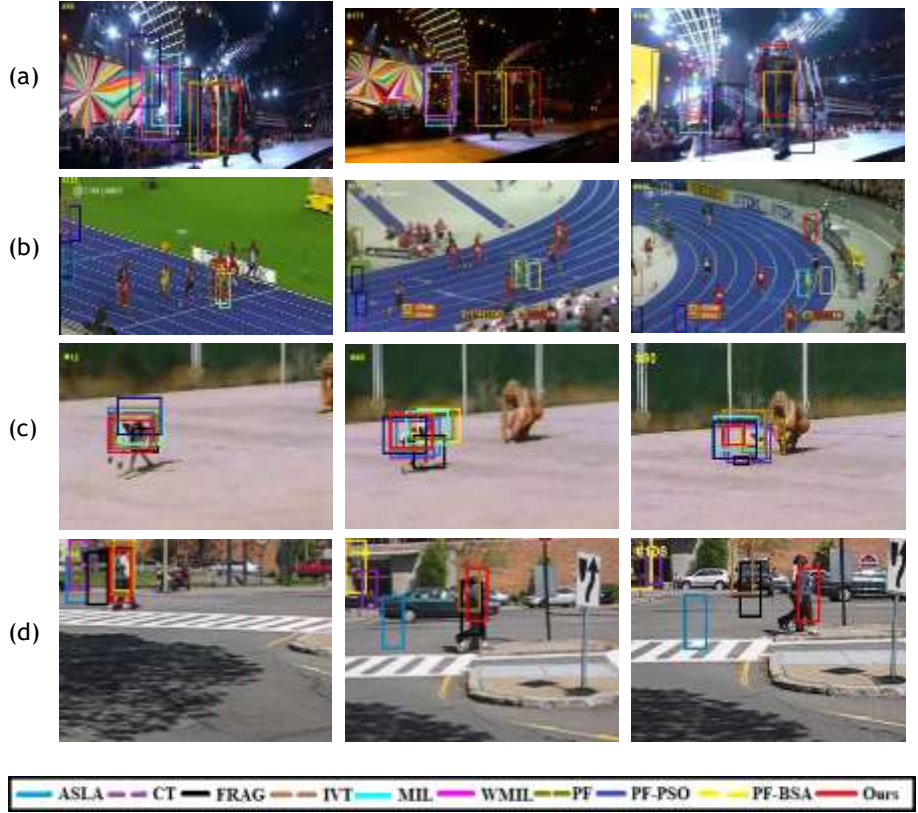

Figure 6: Critical frames with Object's deformation and rotation for video sequences: a) *Singer2* b) *Bolt* c) *Dog* d) *Couple*

| Video Sequence | ASLA | CT | FRAG | IVT | MIL | WMIL | PF | PF-PSO | PF-BSA | Ours |
|---|---|---|---|---|---|---|---|---|---|---|
| *CarScale* | 0.460 | 0.432 | 0.471 | 0.002 | 0.172 | 0.474 | 0.176 | 0.353 | 0.348 | 0.786 |
| *Dog* | 0.500 | 0.492 | - | 0.109 | 0.481 | 0.378 | 0.202 | 0.487 | 0.244 | 0.610 |
| *Human8* | 0.130 | 0.063 | 0.218 | 0.049 | 0.071 | 0.087 | 0.379 | 0.156 | 0.384 | 0.733 |
| *Singer2* | 0.053 | 0.138 | 0.246 | 0.049 | 0.048 | 0.064 | 0.395 | 0.331 | 0.417 | 0.867 |
| *crowds* | 0.071 | 0.004 | - | 0.075 | 0.059 | 0.005 | 0.056 | 0.602 | 0.280 | 0.777 |
| *Jogging2* | 0.133 | 0.092 | 0.569 | 0.158 | 0.138 | 0.133 | 0.177 | 0.651 | 0.483 | 0.852 |
| *Subway* | 0.194 | 0.699 | 0.757 | 0.183 | 0.801 | 0.189 | 0.321 | 0.605 | 0.572 | 0.687 |
| *Bolt* | 0.020 | 0.028 | 0.019 | 0.008 | 0.013 | 0.015 | 0.168 | 0.207 | 0.129 | 0.709 |
| *Walking* | 0.069 | 0.643 | 0.684 | 0.874 | 0.685 | 0.659 | 0.037 | 0.493 | 0.091 | 0.498 |
| *Mountainbike* | 0.633 | 0.222 | 0.149 | 0.840 | 0.821 | 0.436 | 0.171 | 0.428 | 0.211 | 0.845 |
| *Crossing* | 0.380 | 0.779 | 0.349 | 0.401 | 0.837 | 0.660 | 0.327 | 0.596 | 0.447 | 0.829 |
| *Couple* | 0.056 | 0.237 | 0.692 | 0.081 | 0.100 | 0.526 | 0.367 | 0.484 | 0.138 | 0.617 |
| **Average** | 0.225 | 0.319 | 0.415 | 0.236 | 0.352 | 0.302 | 0.231 | 0.449 | 0.312 | 0.734 |

Table 3: F-Measure. The best results are shown in red, blue and green respectively.

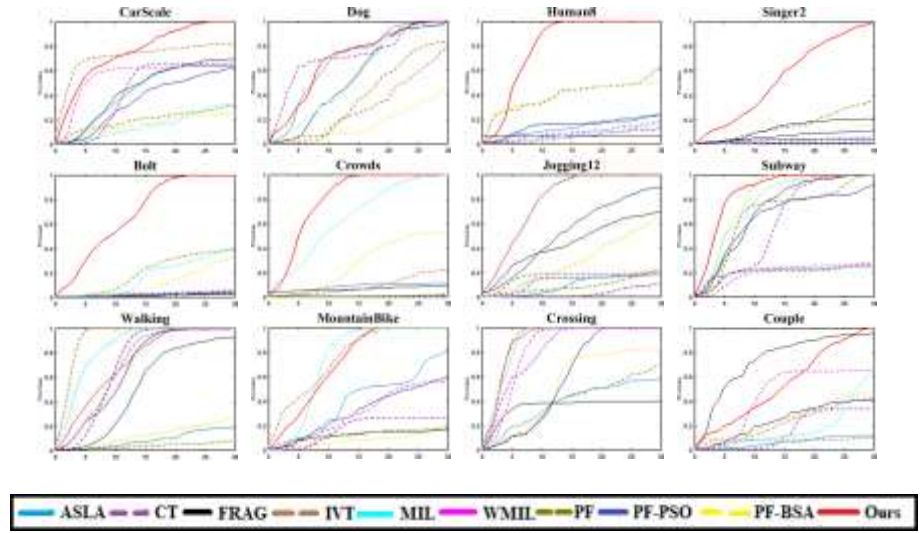

Figure 7: Precision Plot *vs* CLE threshold (in pixels) under various visual tracking challenges:
i) Illumination and Scale variations (*CarScale*, *Dog*, *Human8*, *Singer2* ) ii) Full and Partial Occlusion (*Bolt*, *Crowds*, *Jogging-2*, *Subway*) iii) Background Clutters (*Walking*, *MountainBike*, *Crossing*, *Couple*) iv) Object's Deformation and Rotation (*Singer2*, *Bolt*, *Dog*, *Couple*)

*4.1.4. Object's Deformation and Rotation*

The sample frames for the challenging video sequences under this challenge are presented in Fig. 6. In *Singer2* video sequence at frames #89 and #132 the target deformed due to its sudden pose variations and subsequent rotation. MIL, CT, WMIL, ASLA, CT and PF-BSA has lost the target. However, ours has shown better results. In Fig. 6(b) critical frames for *Bolt* sequences are shown. Non-rigid deformation occurs in the target due to its in-plane rotation at frames #274 and #340. Most of the trackers are unable to handle this deformation. Only our tracker manages to handle the target's deformation. The sample frames for *Dog* sequence are shown in fig. 6(c). At frames #12 and #46 the target has deformation due its fast motion and rotation. MIL, FRAG, PF and PF-BSA are slightly drifted from the target whilst ours, ASLA and CT perform efficiently. *Couple* whose sample frames are presented in Fig. 5(d). At frame #100 ASLA, CT and PF-BSA fail to keep track of the target as deformation occurs due to sudden camera movement. Ours, FRAG and IVT keep track of the target. However, at frame #108 only our tracker is able to locate the target. It is mainly due to the rotational component which is incorporated in the state vector of the multi component state model. In addition, the low performing particles due to object's deformation are suppressed by the multicue fuzzy fusion model to cater this challenge.

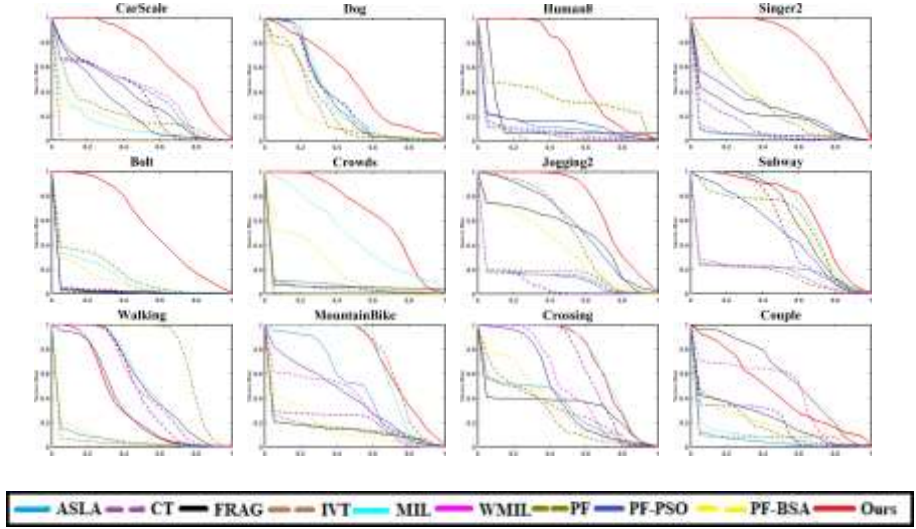

Figure 8: Success *vs* Percent Overlap Threshold (%) Plot under various visual tracking challenges: i) Illumination and Scale variations (*CarScale*, *Dog*, *Human8*, *Singer2* ) ii) Full and Partial Occlusion (*Bolt*, *Crowds*, *Jogging-2*, *Subway*) iii) Background Clutters (*Walking*, *MountainBike*, *Crossing*, *Couple*) iv) Object's Deformation and Rotation (*Singer2*, *Bolt*, *Dog*, *Couple*)

*4.2. Quantitative analysis*

The quantitative performance of proposed tracker is analyzed by considering multiple evaluation metrics which includes Centre location error(CLE), F-Measure [40] and plots as: Precision vs Location error threshold and Success rate vs overlap threshold [41]. CLE is calculated as the distance between tracker bounding box and the center of the ground truth. F-Measure is defined as $\frac{2 \times pr \times rl}{pr+rl}$
Where $pr$ is precision and calculated as $(B_R \cap B_G)/B_R$ ; $rl$ is recall and given as: $\frac{B_R \cap B_G}{B_G}$. $B_R$ is tracker bounding box and $B_G$ ground truth bounding box. Precision plot illustrates the mean precision at various location error thresholds (in pixels). The Success plot represents the percent success over various overlap thresholds (in %).

We have achieved an average CLE of 7.98 and average F-Measure of 0.734

when evaluated on 12 challenging video sequences taken from benchmark datasets. The mean CLE and mean F-Measure are tabulated in Table 2 and Table 3 respectively. Fig.(7) represents the percentage overlap of successful frames from a given threshold from the ground truth. Fig.(8) illustrates the success rate over overlap threshold. As shown in Fig. (7), for 20 location error threshold (in pixels) we achieve an average precision of 0.90 for illumination and scale variations, 0.98 for full or partial occlusion, 0.91 for background clutters and 0.82 for object's deformation and rotation. As depicted in Fig.(8), at 0.2 overlap threshold (in %) our tracker manages average success rate 0.99 for illumination and scale variations as well as for full or partial occlusion, 0.97 for background clutters and 0.95 for object's deformation and rotation. High values of precision and success rate reveals that our tracker can handle various dynamic environment challenges. On the other hand, generative tracker such as ASLA [35] are not robust against full or partial occlusion as its appearance model do not consider background information. Discriminative trackers viz. CT[36] and MIL[37] fail to keep track of target in background clutter and illumination variation. WMIL[39] is considered random Haar like features for object representation and hence, not able to tackle sudden illumination and scaling variations. IVT[38] performance is limited by the large number of parameters used for developing appearance model of the target. PF based trackers such as PF[6], PF-PSO[15], PF-BSA[27] are not efficient enough to address dynamic challenges, mainly due to limited solution to sample degeneracy and impoverishment problem. Our tracker has exploited CSA as a resampling technique which ensures the diversity in the search space with two parameters: awareness probability and flight length. The outlier detection mechanism helps in identifying the low performing particles. The complexity of the proposed resampling technique is reduced as it resampled only the low performing particles detected by the outlier. In addition, the proposed multicue fuzzy based fusion model not only boost the high performing particles but also, suppresses the low performing particles. Hence, achieve high accuracy and robustness against dynamic challenges is achieved.

## 5. Conclusion

In this paper, we propose a robust tracker based on multicue particle filter framework which comprises of adaptive multicue fuzzy based fusion model and a novel resampling method to overcome particle degeneracy. Complementary multicue viz. RGB color and LBP texture are considered to enhance the tracker accuracy under illumination variation, occlusion, background clutters, object's deformation and rotation. The adaptive mulitcue fuzzy fusion model either boosts or suppresses the particles in order to ensure clear decision boundary between the unimportant particles and the significant particles. Nevertheless, an outlier detection mechanism is also proposed to classify the unimportant particles from the significant one. Proposed tracker is computationally efficient as only the unimportant particles are resampled by the proposed resampling technique. Quick adaptation of tracker is ensured through online estimation of transductive reliability of each cue. Qualitative and quantitative analysis on 12 challenging video sequences from benchmark infer that the proposed tracker perform favorable against the state-of-the-art trackers.

In future, we look forward to integrate more multimodal features in the proposed tracking framework. Inclusion of user defined source importance in the proposed fusion model is another direction for future undertaking. We would also like to extend our work for tracking multiple people in video sequences. Online model for guiding particles under uncertain conditions can also be investigated.